\documentclass[conference]{IEEEtran}
\usepackage[utf8]{inputenc}
\IEEEoverridecommandlockouts
\usepackage{cite}
\usepackage{amsmath,amssymb,amsfonts}
\usepackage{algorithmic}
\usepackage{graphicx}
\usepackage{textcomp}
\usepackage{xcolor}
\usepackage{url} 
\usepackage{multirow}
\usepackage{booktabs}
\def\BibTeX{{\rm B\kern-.05em{\sc i\kern-.025em b}\kern-.08em
    T\kern-.1667em\lower.7ex\hbox{E}\kern-.125emX}}
\usepackage{fancyhdr}
\definecolor{darkgreen}{rgb}{0.0, 0.5, 0.0}

\begin{document}

\cfoot{\small{\textrm{CIIT 2025 22nd International Conference on Informatics and Information Technologies (CIIT)}}}
\title{Accelerating IoV Intrusion Detection: Benchmarking GPU-Accelerated vs CPU-Based ML Libraries\\
}

\makeatletter
\newcommand{\linebreakand}{%
 \end{@IEEEauthorhalign}
  \hfill\mbox{}\par
  \mbox{}\hfill\begin{@IEEEauthorhalign}
  
}

\author{
\IEEEauthorblockN{1\textsuperscript{st} Furkan Çolhak}
\IEEEauthorblockA{
\textit{CCIP, Center for Cyber Security and}\\
\textit{Critical Infrastructure Protection,}\\
\textit{Kadir Has University}\\
Istanbul, Turkey \\
furkancolhak@stu.khas.edu.tr
}
\and
\IEEEauthorblockN{2\textsuperscript{nd} Hasan Coşkun}
\IEEEauthorblockA{
\textit{CCIP, Center for Cyber Security and}\\
\textit{Critical Infrastructure Protection,}\\
\textit{Kadir Has University}\\
Istanbul, Turkey \\
hasan.coskun@stu.khas.edu.tr
}

\and
\IEEEauthorblockN{3\textsuperscript{rd} Tsafac Nkombong Regine Cyrille}
\IEEEauthorblockA{
\textit{CyberMACS,}\\
\textit{Kadir Has University}\\
Istanbul, Turkey \\
tsafacnkombong.cyrille@stu.khas.edu.tr
}
\linebreakand
\IEEEauthorblockN{4\textsuperscript{th} Tedi Hoxa}
\IEEEauthorblockA{
\textit{CyberMACS,}\\
\textit{Kadir Has University}\\
Istanbul, Turkey \\
tedihoxa@stu.khas.edu.tr
}
\and
    
\IEEEauthorblockN{5\textsuperscript{th} Mert İlhan Ecevit}
\IEEEauthorblockA{
\textit{CCIP, Center for Cyber Security and}\\
\textit{Critical Infrastructure Protection,}\\
\textit{Kadir Has University}\\
Istanbul, Turkey \\
mertilhan.ecevit@khas.edu.tr
} 
\and
\IEEEauthorblockN{6\textsuperscript{th} Mehmet Nafiz Aydın}
\IEEEauthorblockA{
\textit{CyberMACS,}\\
\textit{Kadir Has University}\\
Istanbul, Turkey \\
mehmet.aydin@khas.edu.tr
}
}

\maketitle

\begin{abstract}
The Internet of Vehicles (IoV) may face challenging cybersecurity attacks that may require sophisticated intrusion detection systems, necessitating a rapid development and response system. This research investigates the performance advantages of GPU-accelerated libraries (cuML) compared to traditional CPU-based implementations (scikit-learn), focusing on the speed and efficiency required for machine learning models used in IoV threat detection environments. The comprehensive evaluations conducted employ four machine learning approaches (Random Forest, KNN, Logistic Regression, XGBoost) across three distinct IoV security datasets (OTIDS, GIDS, CICIoV2024). Our findings demonstrate that GPU-accelerated implementations dramatically improved computational efficiency, with training times reduced by a factor of up to 159 and prediction speeds accelerated by up to 95 times compared to traditional CPU processing, all while preserving detection accuracy. This remarkable performance breakthrough empowers researchers and security specialists to harness GPU acceleration for creating faster, more effective threat detection systems that meet the urgent real-time security demands of today's connected vehicle networks.
\end{abstract}


\section{Introduction}
In recent years, the Internet of Things (IoT) has gained widespread adoption, connecting devices to improve efficiency in sectors such as transportation, healthcare and energy \cite{nivzetic2020internet}. A key extension of IoT is the Internet of Vehicles (IoV), which integrates vehicles into smart automotive environments to support real-time monitoring, predictive maintenance, and autonomous driving \cite{xu2017internet}. Despite the fact that IoV brings significant benefits, it also faces some important cybersecurity challenges\cite{pinto2024}.

Since IoV vehicles can directly affect the safety of life and property, it is important for IoV vehicles to implement strong security measures, with Intrusion Detection Systems (IDS) being essential for detecting and mitigating cyber threats. A notable example of these risks occurred in 2015, when security researchers demonstrated a critical vulnerability (CVE-2015-5611) in Fiat Chrysler’s Uconnect system, affecting vehicles manufactured between 2013 and 2015. The researchers remotely compromised a 2014 Jeep Cherokee through its entertainment system, gaining access to critical vehicle controls, including steering, brakes, and transmission\cite{miller2015remote}. Exploitation of this vulnerability may allow an unauthorized user to take remote control of an affected vehicle, but the attack requires access. This incident led to the recall of 1.4 million vehicles and highlighted the severe consequences of insufficient security measures in connected vehicles. More recently, a Volkswagen data breach exposed sensitive information about 800,000 owners of electric vehicles, including detailed GPS locations and usage patterns \cite{spiegel}. These incidents emphasize the crucial importance of adopting robust and innovative IoV cybersecurity measures and effective intrusion detection systems (IDS).

\begin{table*}[t]
    \caption{Overview of Intrusion Detection Studies in Vehicle Systems (Condensed Models)}
    \centering
    \renewcommand{\arraystretch}{1.1} 
    \setlength{\tabcolsep}{4pt}
    \begin{tabular}{|p{1.5cm}|p{10cm}|p{2cm}|p{1.5cm}|p{2cm}|} 
        \hline
        \textbf{Author(s)} & \textbf{Study Summary (incl. Comp. Time/Efficiency Focus)} & \textbf{ML Models} & \textbf{Libraries} & \textbf{Dataset} \\
        \hline
        Purohit and Govindarasu (2022) \cite{9850292} 
        & The study optimizes training and execution time by using efficient tree-based algorithms like Decision Tree, Random Forest, and XGBoost, combined with a rule-based system, achieving high accuracy and low latency for real-time CAN bus intrusion detection.
        & DT, RF, XGB  
        & Keras, Pandas 
        & OTIDS\\
        \hline
        Saber and Mazri (2023) \cite{10322964} 
        &  The study emphasizes the importance of efficient feature selection and classifier optimization to enhance detection accuracy and reduce computational overhead.
        & KNN, DT, NB, SVM, LR, RF  
        & NumPy, SciPy, Scikit-learn
        & GIDS \\
        \hline
        Liu and Fan (2023) \cite{10258136} 
        & The study employs Recursive Feature Elimination (RFE) to reduce data dimensions, improving detection efficiency. Decision Trees and Multi-Layer Perceptron (MLP) are used for classification and optimizing execution speed.
        & DT, MLP 
        & \textit Not explicitly mentioned
        & CICIDS17, GIDS \\
        \hline
        T.-T.-H. Le et al. (2023) \cite{le2023enhancing}& Proposes XGBoost with SHAP explanations for IDS on imbalanced CAN data. Focuses on binary/multiclass accuracy and interpretability. Computational time/efficiency is not the primary focus, but efficiency is inherent in choosing XGBoost over some DL methods. & XGB& Scikit-learn (implied) & GIDS \\
        \hline
        Limouchi \& Chan (2023) \cite{limouchi2023optimized}   & Optimized ML (LightGBM, ET, XGBoost) for IoV IDS. Addresses imbalance (SMOTE), uses Correlation-based Feature Selection (FS), and Bayesian Optimization (BO) for hyperparameters. Focuses on performance enhancement and reducing computational cost via FS. & LGBM, ET, XGB & Pandas, Matplotlib, Seaborn, Scikit-learn & CICDDoS2019 \\
        \hline
        Kumar \& Singh (2023) \cite{kumar2023enhancing} & Uses tree-based ML (DT, RF, ET, XGBoost) and Stacking ensemble for AV IDS. Addresses imbalance (SMOTE) and uses feature selection. Aims for high detection rates at low processing costs, highlighting computational efficiency as a consideration. & DT, RF, ET, XGB & Not explicitly mentioned & CICIDS2017, GIDS \\
        \hline
        Gou et al.(2023) \cite{Gou2023}& Proposes an Adaptive Tree-Based Ensemble Network (ATBEN) with stacked ML models for multiclass IDS in IoV. Addresses imbalance (SMOTE + RUS) and uses ML-based feature selection (FS). Aims for efficient multiclass classification, mentions reducing computational overhead. & ATBEN (RF, ET, XGB, LGBM etc.) & PyTorch & CICIDS2017, GIDS \\
        \hline
        Kaushik et al. (2024) \cite{10541401} 
        & Optimizes training and execution time using feature selection with Mutual Information and Correlation, achieving significant reductions in computational time across various classifiers.
        & DT, LR, LDA, NB  
        & Scikit-learn (implied) 
        & CICIDS18 \\
        \hline
        Korium et al. (2024) \cite{KORIUM2024103330} 
        & \textbf Utilizes ML models with Z-score normalization and regression-based feature selection to enhance model efficiency, achieving high accuracy and optimized execution and detection times.
        & RF, XGB, CatBoost, LGBM 
        &Scikit-learn 
        & CICIDS17, CSEIDS18, CICDDoS19 \\
        \hline
        
    \end{tabular}
\end{table*}

The increasing sophistication of IoV attacks, combined with the massive scale and complexity of vehicle networks, creates unique challenges for intrusion detection systems, particularly in the context of real-time threat detection. Although traditional IDS approaches have proven effective in conventional networks, vehicular networks' real-time requirements and complex nature demand more efficient computational solutions. Machine learning-based IDS approaches have shown promising results, with techniques such as Extreme Gradient Boosting and tree-based models demonstrating high detection accuracy \cite{Bajpai2023,Rani2023}. However, the computational demands of these ML models present a significant challenge in IoV environments, where real-time detection is crucial \cite{10541401}. The choice between GPU-accelerated and CPU-based implementations can significantly impact both detection accuracy and response time, yet there is limited research comparing their performance in IoV contexts.

This paper addresses this challenge by investigating the performance of machine learning models in IoV environments, focusing on the computational efficiency between GPU-accelerated libraries (cuML) and CPU-based implementations (scikit-learn). Using three distinct IoV datasets: CAN-intrusion dataset (OTIDS), Car-hacking dataset (GIDS), and CICIoV2024. This work evaluates the models using key metrics, including accuracy, F1 score, training time, prediction time and speedup metrics.

\noindent Our research has made contributions that can be outlined as follows:
\begin{itemize} 
    \item This study systematically compares GPU-accelerated (cuML) versus CPU-based (Scikit-learn) machine learning for IoV tasks, analyzing both performance (accuracy, F1 score) and computational efficiency (training/prediction times, speedup factors).
    \item The evaluation uses three diverse IoV datasets (OTIDS, GIDS, and CICIoV2024) to comprehensively and realistically assess model suitability in different scenarios. 
\end{itemize}

\begin{table*}[t]
\caption{IoV IDS Dataset Properties}
\centering
\renewcommand{\arraystretch}{1.1}
\setlength{\tabcolsep}{5pt}
\begin{tabular}{|p{5cm}|p{1.5cm}|p{2.5cm}|p{2.5cm}|p{2.5cm}|p{1.5cm}|}
\hline
\textbf{Dataset} & \textbf{Year} & \textbf{Sample Count} & \textbf{Class} & \textbf{Count} & \textbf{\%} \\
\hline
\multirow{4}{*}{CAN-intrusion-dataset (OTIDS) \cite{lee2017otids}} 
& \multirow{4}{*}{2017} & \multirow{4}{*}{4,613,439}  
& Benign & 2,369,398 & 51.35 \\ 
& & & DoS & 656,579 & 14.23 \\ 
& & & Fuzzy & 591,990 & 12.83 \\ 
& & & Impersonation & 995,472 & 21.57 \\ 
\hline

\multirow{5}{*}{Car-Hacking Dataset (GIDS) \cite{seo2018gids}} 
& \multirow{5}{*}{2018} & \multirow{5}{*}{14,427,180}  
& Benign & 2,056,938 & 6.85 \\ 
& & & RPM & 4,621,702 & 32.03 \\ 
& & & DoS & 3,665,771 & 25.40 \\ 
& & & Fuzzy & 2,056,938 & 14.25 \\ 
& & & Gear & 3,093,898 & 21.44 \\ 
\hline

\multirow{6}{*}{CICIoV2024 \cite{pinto2024}} 
& \multirow{6}{*}{2024} & \multirow{6}{*}{1,408,219}  
& Benign & 1,223,737 & 86.89 \\ 
& & & DoS & 74,663 & 5.30 \\ 
& & & RPM & 54,900 & 3.89 \\ 
& & & Speed & 24,951 & 1.77 \\ 
& & & Steering Wheel & 19,977 & 1.41 \\ 
& & & Gas & 9,991 & 0.70 \\ 
\hline
\end{tabular}
\label{table:compact_dataset}
\end{table*}

\section{Literature Review}
\label{sec2}
\subsection{Research Methodology}
In conducting a comprehensive review of the literature on Intrusion Detection Systems (IDS) within the Internet of Vehicles (IoV) framework, this work utilized multiple databases to ensure a broad coverage of related studies. 

These databases included IEEE Xplore, Scopus, Web of Science, and ACM Digital Library. The queries were constructed around key elements of our research focus, covering IoV concepts, security and IDS, machine learning techniques, data challenges, and specific datasets. Additionally, the search was restricted to journal and conference papers published after 2022.

The following queries were employed:

\begin{itemize} 
    \item \textbf{Query 1:} \texttt{"Internet of Vehicles" OR "IoV" OR "Vehicular Network" OR "Connected Vehicles"}
    \item \textbf{Query 2:} \texttt{"Intrusion Detection System" OR "IDS" OR "Cybersecurity" OR "Attack Detection"} 
    \item \textbf{Query 3:} \texttt{"Machine Learning" OR "ML"}
    \item \textbf{Query 4:} \texttt{"Training Time" OR "Execution Time" OR "Computational Efficiency"} 
\end{itemize}

The initial search produced many results: 16,218 publications across all databases for the broadest query set. Applying the subsequent filters and combining queries gradually narrowed this number. Combining Query 1 and Query 2 reduced the results to 1,206 studies. Further refinement by adding Query 3 led to 348 publications. The most specific search, incorporating Queries 1, 2, 3, and 4, resulted in only 24 studies. After a thorough review to eliminate duplicates and exclude studies not focused on IoV or IDS, only seven studies were directly relevant for detailed analysis. These seven papers were carefully selected based on their focus on IoV security, the use of machine learning algorithms and libraries, and their discussion of computational aspects such as training and execution time.

\subsection{Review of the Studies}
The increasing connectivity of vehicles within the Internet of Vehicles (IoV) paradigm has significantly expanded the attack surface, making robust Intrusion Detection Systems (IDS) essential for ensuring safety and security. Recent research has extensively explored Machine Learning (ML) techniques to develop effective IDSs capable of identifying malicious activities within vehicular networks, including both Controller Area Network (CAN) bus communications and external IoV interactions.

A review of contemporary studies reveals a diverse range of ML approaches. Tree-based algorithms, such as Decision Trees (DT), Random Forest (RF), XGBoost (XGB), LightGBM (LGBM), and Extra Trees (ET), are frequently employed due to their efficiency and interpretability, as demonstrated by Purohit and Govindarasu (2022), Limouchi \& Chan (2023), Kumar \& Singh (2023), Korium et al. (2024), and Gou et al. (2023). Other classic classifiers like K-Nearest Neighbors (KNN), Support Vector Machines (SVM), Logistic Regression (LR), Naive Bayes (NB), and Linear Discriminant Analysis (LDA) also find application (Saber and Mazri, 2023; Alshathri et al., 2024; Kaushik et al., 2024). More complex architectures, including Multi-Layer Perceptrons (MLP) (Liu and Fan, 2023), Graph Neural Networks (GNN) (He et al., 2023), and sophisticated ensemble methods like Stacking (Kumar \& Singh, 2023) and adaptive tree-based networks (ATBEN) (Gou et al., 2023), are also being investigated. Interpretability is sometimes addressed using methods like SHAP (T.-T.-H. Le et al., 2023).

Computational efficiency is a critical consideration for real-time vehicular IDS. Several studies explicitly aim to optimize training and execution times. Purohit and Govindarasu (2022) focused on low latency using efficient tree models. Feature selection is frequently cited as a means to reduce computational load (Limouchi \& Chan, 2023; Kaushik et al., 2024; Liu and Fan, 2023). Korium et al. (2024) specifically targeted optimized execution and detection times, while Kumar \& Singh (2023) aimed for high detection rates at low processing costs. These optimizations are typically performed using standard CPU-based libraries, predominantly Scikit-learn, alongside libraries like Pandas and NumPy and specific ML framework libraries like Keras or PyTorch.

While these studies advance ML-based IDS for IoV, focusing on accuracy, imbalance, and feature dimensionality often within CPU constraints, there remains a need to explore alternative computational paradigms for further efficiency gains. The potential benefits of GPU acceleration, leveraging libraries specifically designed for accelerating data science workflows like cuML, have not been systematically compared against traditional CPU-based approaches (like Scikit-learn) for these specific IoV intrusion detection tasks.

Therefore, this study aims to bridge this gap by systematically comparing the performance (accuracy, F1 score) and computational efficiency (training time, prediction time, speedup factors) of GPU-accelerated (cuML) versus CPU-based (Scikit-learn) ML implementations. By utilizing diverse IoV datasets (OTIDS, GIDS, and CICIoV2024), this work provides a comprehensive assessment of the practical benefits and trade-offs of GPU acceleration for enhancing real-time intrusion detection capabilities in vehicular networks.

\section{Datasets} \label{sec3} 
The scarcity of IoV-based attack datasets masquerades as a primary challenge in security research. Due to this scarcity, researchers frequently depend on IoT-based datasets like CICIDS2017\cite{Sharafaldin2018} and CICDDoS2019\cite{8888419}, which contain traffic data for attack types such as brute force, DoS, and botnets. Whereas the datasets offer crucial insights for establishing baseline methodologies and evaluation frameworks, they break down address IoV-specific challenges, including vehicular network protocols and domain-specific vulnerabilities. To get meaningful and applicable results, working with domain-specific datasets is essential. This study focuses on IoV-specific datasets, ensuring intrusion detection research stays relevant and effective in addressing real-world security challenges\cite{pinto2024}.

Among the three datasets which are analyzed, stand out for their suitability:  OTIDS \cite{lee2017otids}, GIDS \cite{seo2018gids}, and CICIoV2024 \cite{pinto2024}. These datasets include real CAN data from various vehicles featuring benign samples and different types of attacks, such as DoS, fuzzing, spoofing, and impersonation, to evaluate intrusion detection systems and security mechanisms for vehicular networks. Further details about these datasets are presented in Table \ref{table:compact_dataset}.

\subsection{Data Preprocessing}
Data representations can vary significantly across datasets. In this research, the focus is on decimal data representation. For datasets that do not support decimal representation natively, conversion is straightforward. Initially, the datasets in this study were provided as separate files. All datasets were merged into one CSV file to simplify data processing and analysis. Notable attributes were retained, such as timestamps, IDs, DLC (Data Length Code), and data bytes (payloads). Also, the data bytes were split into individual bits and converted from hexadecimal to decimal format for better usability.

Some anomalies were identified in the OTIDS and GIDS datasets in the preprocessing phase. Irrelevant rows were removed, and missing values (NaN) were replaced with -1 to indicate missing data in a manner compatible with machine learning algorithms. Subsequently, timestamps and IDs were excluded from the datasets, as they were collected in a simulated environment and could inadvertently provide insights to machine learning models. Additionally, the DLC (Data Length Code) column was removed.

The final datasets contain only bit columns (DATA\_0 to DATA\_7) and the corresponding label. Data distributions are presented in Table \ref{table:compact_dataset}. Finally, object labels were numerically encoded using LabelEncoder, and features were standardized using StandardScaler for model training.

\section{Model Selection and Implementation}
\subsection{Machine Learning Algorithms}
The following machine learning models were selected based on their effectiveness in intrusion detection tasks and relevance to IoV datasets.

\textbf{Logistic Regression (LR)} is a basic statistical model for classification, estimating probabilities to assign input data to categories. This study adapted it for multiclass classification using the OnevsOneClassifier, making it proper for detecting IoV attack types. While basic and efficient for smaller datasets, its performance degrades with increasing data size because of its linear nature, making it less competitive in complex IoV scenarios~\cite{Sutanto2024}.

\textbf{K-Nearest Neighbor (KNN)} classifies data points based on their similarity to labeled neighbors, making it a simple yet flexible choice for multiclass problems. Although effective for smaller datasets, KNN deals with large datasets as the computational cost significantly increases the number of data points. In real-time IoV applications, this results in less scalability and slower forecasts~\cite{Sutanto2024}.

\textbf{Random Forest (RF)} is an ensemble method that merges multiple decision trees, offering robustness to noise and adaptability to complex datasets. Its ability to manage high-dimensional data makes it ideal for IoV systems. On the other hand, its training times increase with data size and the number of trees, which can challenge its practicality in time-sensitive scenarios.~\cite{Sutanto2024}.

\textbf{Extreme Gradient Boosting (XGBoost)} is a powerful enhancement of gradient boosting that sequentially builds decision trees to correct prior errors. Its ability to handle missing values and deliver high performance on structured datasets makes it particularly valuable for IoV systems. Despite these strengths, its computational complexity and reliance on careful hyperparameter tuning pose challenges for large datasets or limited computational resources~\cite{price2023}.
\begin{table*}[t]
\caption{Model Performance Comparison Across Different Datasets (Inline Speedup)} 
\centering
\renewcommand{\arraystretch}{1} 
\setlength{\tabcolsep}{14pt} 

\small
\begin{tabular}{@{}l l l r r l l@{}} 
\toprule 
\textbf{Dataset} & \textbf{Model} & \textbf{Library} & \textbf{Accuracy (\%)} & \textbf{F1-Score (\%)} & \textbf{Train Time (s)} & \textbf{Pred. Time (s)} \\ 
\midrule 

\multirow{8}{*}{OTIDS} 
 & \multirow{2}{*}{RFC} 
    & Scikit-learn & 82.37 & 82.17 & 279.28 & 5.49 \\  
 &  & cuML         & 77.23 & 77.61 & 8.00 \ \ {\footnotesize\color{darkgreen}($\downarrow$ 34.9x)} & 0.28 \ \ {\footnotesize\color{darkgreen}($\downarrow$ 19.6x)} \\ 
\cmidrule(lr){2-7} 
 & \multirow{2}{*}{kNN} 
    & Scikit-learn & 81.76 & 81.57 & 11.47 & 48.23 \\  
 &  & cuML         & 81.75 & 81.56 & 0.11 \ \ {\footnotesize\color{darkgreen}($\downarrow$ 104.3x)} & 1.43 \ \ {\footnotesize\color{darkgreen}($\downarrow$ 33.7x)} \\  
\cmidrule(lr){2-7} 
 & \multirow{2}{*}{LR} 
    & Scikit-learn & 54.12 & 40.23 & 7.72 & 0.08 \\  
 &  & cuML         & 54.12 & 40.24 & 1.15 \ \ {\footnotesize\color{darkgreen}($\downarrow$ 6.7x)} & 0.04 \ \ {\footnotesize\color{darkgreen}($\downarrow$ 2x)} \\ 
\cmidrule(lr){2-7} 
 & \multirow{2}{*}{XGB} 
    & Scikit-learn & 79.60 & 79.07 & 140.08 & 0.71 \\  
 &  & cuML         & 79.60 & 79.07 & 26.34 \ \ {\footnotesize\color{darkgreen}($\downarrow$ 5.3x)} & 0.22 \ \ {\footnotesize\color{darkgreen}($\downarrow$ 3.2x)} \\  
\midrule 

\multirow{8}{*}{GIDS} 
 & \multirow{2}{*}{RFC} 
    & Scikit-learn & 60.44 & 58.85 & 1603.99 & 46.19 \\  
 &  & cuML         & 57.20 & 55.28 & 36.01 \ \ {\footnotesize\color{darkgreen}($\downarrow$ 44.5x)} & 0.48 \ \ {\footnotesize\color{darkgreen}($\downarrow$ 96.0x)} \\  
\cmidrule(lr){2-7} 
 & \multirow{2}{*}{kNN} 
    & Scikit-learn & 50.77 & 51.49 & 44.56 & 27793.30 \\ 
 &  & cuML         & 49.82 & 50.18 & 0.28 \ \ {\footnotesize\color{darkgreen}($\downarrow$ 159.1x)} & 494.46 \ \ {\footnotesize\color{darkgreen}($\downarrow$ 56.2x)} \\ 
\cmidrule(lr){2-7} 
 & \multirow{2}{*}{LR} 
    & Scikit-learn & 38.09 & 33.68 & 25.58 & 0.27 \\  
 &  & cuML         & 37.96 & 33.56 & 1.81 \ \ {\footnotesize\color{darkgreen}($\downarrow$ 14.1x)} & 0.07 \ \ {\footnotesize\color{darkgreen}($\downarrow$ 3.9x)} \\  
\cmidrule(lr){2-7} 
 & \multirow{2}{*}{XGB} 
    & Scikit-learn & 60.39 & 58.61 & 214.42 & 2.14 \\  
 &  & cuML         & 60.43 & 58.66 & 24.97 \ \ {\footnotesize\color{darkgreen}($\downarrow$ 8.6x)} & 0.51 \ \ {\footnotesize\color{darkgreen}($\downarrow$ 4.2x)} \\  
\midrule 

\multirow{8}{*}{CICIoV2024} 
 & \multirow{2}{*}{RFC} 
    & Scikit-learn & 99.64 & 99.63 & 49.45 & 0.97 \\  
 &  & cuML         & 99.29 & 99.24 & 4.33 \ \ {\footnotesize\color{darkgreen}($\downarrow$ 11.4x)} & 0.08 \ \ {\footnotesize\color{darkgreen}($\downarrow$ 12.1x)} \\  
\cmidrule(lr){2-7} 
 & \multirow{2}{*}{kNN} 
    & Scikit-learn & 99.64 & 99.65 & 2.25 & 382.46 \\  
 &  & cuML         & 99.64 & 99.63 & 0.04 \ \ {\footnotesize\color{darkgreen}($\downarrow$ 56.3x)} & 6.59 \ \ {\footnotesize\color{darkgreen}($\downarrow$ 58.0x)} \\  
\cmidrule(lr){2-7}
 & \multirow{2}{*}{LR} 
    & Scikit-learn & 87.47 & 84.14 & 19.87 & 0.02 \\  
 &  & cuML         & 87.48 & 84.15 & 6.31 \ \ {\footnotesize\color{darkgreen}($\downarrow$ 3.1x)} & 0.02 \ \ {\footnotesize\color{gray}(1.0x)} \\ 
\cmidrule(lr){2-7} 
 & \multirow{2}{*}{XGB} 
    & Scikit-learn & 99.64 & 99.65 & 134.17 & 0.12 \\  
 &  & cuML         & 99.64 & 99.65 & 14.72 \ \ {\footnotesize\color{darkgreen}($\downarrow$ 9.1x)} & 0.11 \ \ {\footnotesize\color{gray}(1.1x)} \\ 

\bottomrule 
\end{tabular}

\vspace{2pt} 
{\footnotesize \textit{Note:} Values in parentheses like {\color{darkgreen}($\downarrow$ Nx)} next to cuML times indicate its speedup factor compared to Scikit-learn for that operation. The $\downarrow$ symbol denotes faster execution (less time).\par}
\label{tab:combined_results_inline_speedup_compact} 
\end{table*}
\subsection{Computational Libraries}
This section provides an overview of the libraries used in this study. Pandas, NumPy, and scikit-learn were used for data preprocessing, numerical operations, and model evaluation.

\textbf{scikit-learn (CPU-based)}:  A widely used library for machine learning, scikit-learn includes algorithms like Logistic Regression, KNN, and Random Forest. It offers a wide range of parameters for tuning models, making it suitable for smaller datasets. However, it is optimized for CPU-based, single-threaded computation.

\textbf{cuML (GPU-based)}: Part of the RAPIDS suite, cuML leverages NVIDIA CUDA for GPU acceleration, speeding up training and prediction for models like Random Forest, KNN, and Gradient Boosted Trees. It is highly efficient for large datasets, significantly reducing training time compared to CPU-based implementations, and supports hyperparameter tuning for enhanced model performance. However, it has fewer tuning parameters than scikit-learn, which may limit flexibility in certain use cases.

XGBoost is a separate, highly optimized library for gradient boosting; while it operates on the CPU by default, it also supports GPU acceleration through the device parameter or gpu\_hist tree method. 
\section{Results and Discussion}\label{sec4}

\subsection{Evaluation Metrics}
Model evaluation centers on \textbf{Accuracy} and \textbf{F1-score} for performance assessment, where the F1-score's balance of precision/recall helps compare libraries robustly. Computational efficiency is measured by \textbf{training} and \textbf{prediction times} (seconds), and the \textbf{speedup} is calculated as scikit-learn time / cuML time to quantify the acceleration provided by cuML.

\subsection{Experimental Setup}
Tests were conducted on a system with an Intel Core i9-10900X CPU (3.70 GHz) and an NVIDIA RTX A4000 GPU (16 GB VRAM, CUDA enabled). Datasets were stratified and randomly split into 80\% training and 20\% testing sets. Parameters are kept consistent across all models for fair comparison.

\subsection{Performance Evaluation}
This study evaluates the performance of Random Forest Classifier (RFC), k-Nearest Neighbors (kNN), Logistic Regression (LR), and eXtreme Gradient Boosting (XGB) models, comparing Scikit-learn (CPU) and cuML (GPU) implementations across the OTIDS, GIDS, and CICIoV2024 datasets.

As detailed in Table \ref{tab:combined_results_inline_speedup_compact}, the primary advantage of cuML was a consistent and significant acceleration in training and prediction times across all datasets and models. Speedups were particularly dramatic for kNN (e.g., over 100x faster training on GIDS and CICIoV2024) and substantial for RFC (e.g., 35x on OTIDS, 45x on GIDS, 11x on CICIoV2024). While cuML's speed often came with slightly lower accuracy/F1 scores for RFC compared to Scikit-learn (particularly on OTIDS and GIDS), performance metrics for kNN, LR, and XGB remained broadly comparable or nearly identical between the two libraries, even achieving high accuracy with cuML on the CICIoV2024 dataset.

These GPU-accelerated speedups are highly beneficial in domains like the Internet of Vehicles (IoV), where efficient model training and prediction are crucial for real-time applications. However, leveraging cuML presents practical challenges. The library currently lacks specific parameters (e.g., class\_weight crucial for imbalanced data), and users may face difficulties with environment setup and updates, especially concerning CUDA compatibility.

Considering these trade-offs, a hybrid approach is recommended: utilize GPU acceleration via cuML for computationally intensive tasks like hyperparameter tuning or feature selection where speed is paramount. Subsequently, the final, optimized model can be trained on a CPU using Scikit-learn. This strategy leverages GPU speed during exploration while ensuring the final model deployment benefits from wider accessibility, lower cost (especially in cloud environments without dedicated GPUs), and potentially more robust parameter options of CPU-based libraries.

\section{Conclusions and Future Work}\label{sec5}
This study investigated the computational efficiency of GPU-accelerated libraries (cuML) compared to CPU-based implementations (scikit-learn) for machine learning models in IoV intrusion detection. Our comprehensive evaluation across three IoV-specific datasets yielded significant insights into the potential of GPU acceleration for security applications.

The results demonstrated substantial performance improvements, with GPU-accelerated models achieving training time reductions ranging from 6.71x to 159.1x and prediction time improvements of 1.09x to 95.98x compared to CPU implementations. Notably, these significant speed improvements were achieved while maintaining comparable accuracy and F1 scores, indicating no compromise in detection reliability. The RFC and KNN models showed impressive enhancements with GPU acceleration, making them promising candidates for real-time IoV security applications. These findings have important implications for IoV security, as faster training and prediction times enable more frequent model updates and quicker responses to emerging threats. However, challenges remain in implementing these solutions in resource-constrained environments and ensuring scalability across different network sizes.

Future research should focus on developing hybrid CPU-GPU architectures, investigating federated learning approaches for distributed networks, and creating standardized benchmarking frameworks for IoV security solutions. Additionally, field testing in actual IoV environments and research into adversarial attack resistance will be crucial for practical implementation.

\section{Acknowledgments}\label{sec7}
This work was supported partially by the European Union in the framework of ERASMUS MUNDUS, Project CyberMACS (Project \#101082683) (\url{https://cybermacs.eu}).\\

\section*{Declaration of Author Contributions} 
Furkan Çolhak and Hasan Coşkun contributed comprehensively to all stages of this research, including the study design, implementation, data analysis, and initial manuscript writing. Tedi Hoxa and Safac Nkombong Regine Cyrille focused on contributing to the writing and refinement of the manuscript. Mert İlhan Ecevit and Mehmet Nafiz Aydın served as supervisors, providing guidance and oversight for the project.  Artificial intelligence tools assisted with grammatical corrections and language refinement during manuscript preparation. All authors have read and agreed to the published version of the manuscript.

\bibliography{sn-bibliography}  

\begin{thebibliography}{10}

\bibitem{nivzetic2020internet}
S.~Ni{\v{z}}eti{\'c}, P.~{\v{S}}oli{\'c}, D.~L.-d.-I. Gonzalez-De, L.~Patrono, {\em et~al.}, ``Internet of things (iot): Opportunities, issues and challenges towards a smart and sustainable future,'' {\em Journal of cleaner production}, vol.~274, p.~122877, 2020.

\bibitem{xu2017internet}
W.~Xu, H.~Zhou, N.~Cheng, F.~Lyu, W.~Shi, J.~Chen, and X.~Shen, ``Internet of vehicles in big data era,'' {\em IEEE/CAA Journal of Automatica Sinica}, vol.~5, no.~1, pp.~19--35, 2017.

\bibitem{pinto2024}
E.~C. Pinto~Neto, H.~Taslimasa, S.~Dadkhah, S.~Iqbal, P.~Xiong, T.~Rahman, and A.~A. Ghorbani, ``Ciciov2024: Advancing realistic ids approaches against dos and spoofing attack in iov can bus,'' {\em Journal of Computer Networks and Communications}, 2024.
\newblock Accessed: 2024-12-30.

\bibitem{miller2015remote}
C.~Miller and C.~Valasek, ``Remote exploitation of an unaltered passenger vehicle,'' in {\em Black Hat USA}, p.~91, 2015.

\bibitem{spiegel}
P.~Beuth, Flüpke, M.~Hoppenstedt, M.~Kreil, M.~Rosenbach, and R.~Wilkin, ``Volkswagen data breach: Conclusions about the lives of people behind the wheel,'' 2024.
\newblock Accessed: 2025-01-05.

\bibitem{9850292}
S.~Purohit and M.~Govindarasu, ``Ml-based anomaly detection for intra-vehicular can-bus networks,'' in {\em 2022 IEEE International Conference on Cyber Security and Resilience (CSR)}, pp.~233--238, 2022.

\bibitem{10322964}
O.~Saber and T.~Mazri, ``In-vehicle intrusion detection based on machine learning,'' in {\em 2023 10th International Conference on Wireless Networks and Mobile Communications (WINCOM)}, pp.~1--6, 2023.

\bibitem{10258136}
J.~Liu and W.~Fan, ``A machine learning-based intrusion detection approach for intelligent connected vehicles,'' in {\em 2023 24st Asia-Pacific Network Operations and Management Symposium (APNOMS)}, pp.~231--234, 2023.

\bibitem{le2023enhancing}
T.-T.-H. Le {\em et~al.}, ``Enhancing intrusion detection and explanations for imbalanced vehicle can network data,'' in {\em Proceedings of the 12th International Symposium on Information and Communication Technology}, 2023.

\bibitem{limouchi2023optimized}
E.~Limouchi and F.~Chan, ``Optimized machine learning-based intrusion detection system for internet of vehicles,'' in {\em 2023 IEEE Symposium Series on Computational Intelligence (SSCI)}, pp.~1151--1157, 2023.

\bibitem{kumar2023enhancing}
A.~B. Kumar and M.~Singh, ``Enhancing intrusion detection in autonomous vehicles using tree-based machine learning techniques,'' in {\em 2023 3rd Asian Conference on Innovation in Technology (ASIANCON)}, pp.~1--7, 2023.

\bibitem{Gou2023}
W.~Gou, H.~Zhang, and R.~Zhang, ``Multi-classification and tree-based ensemble network for the intrusion detection system in the internet of vehicles,'' {\em Sensors}, vol.~23, no.~21, p.~8788, 2023.

\bibitem{10541401}
S.~Kaushik, A.~Bhardwaj, and S.~S.~M. Rahman, ``Micord-ids: A hybrid learning system for intrusion detection system for the internet of vehicles,'' in {\em 2024 7th International Conference on Information and Computer Technologies (ICICT)}, pp.~485--492, 2024.

\bibitem{KORIUM2024103330}
M.~S. Korium, M.~Saber, A.~Beattie, A.~Narayanan, S.~Sahoo, and P.~H. Nardelli, ``Intrusion detection system for cyberattacks in the internet of vehicles environment,'' {\em Ad Hoc Networks}, vol.~153, p.~103330, 2024.

\bibitem{Bajpai2023}
S.~Bajpai, K.~Sharma, and B.~K. Chaurasia, ``Intrusion detection for internet of vehicles using machine learning,'' in {\em 2023 14th International Conference on Computing Communication and Networking Technologies (ICCCNT)}, IEEE, 2023.

\bibitem{Rani2023}
P.~Rani and R.~Sharma, ``Intelligent transportation system for internet of vehicles based vehicular networks for smart cities,'' {\em Computers and Electrical Engineering}, vol.~105, p.~108543, 2023.

\bibitem{lee2017otids}
H.~Lee, S.~H. Jeong, and H.~K. Kim, ``Otids: A novel intrusion detection system for in-vehicle network by using remote frame,'' in {\em 2017 15th Annual Conference on Privacy, Security and Trust (PST)}, pp.~57--5709, IEEE, 2017.

\bibitem{seo2018gids}
E.~Seo, H.~M. Song, and H.~K. Kim, ``Gids: Gan based intrusion detection system for in-vehicle network,'' in {\em 2018 16th annual conference on privacy, security and trust (PST)}, pp.~1--6, IEEE, 2018.

\bibitem{Sharafaldin2018}
I.~Sharafaldin, A.~H. Lashkari, and A.~A. Ghorbani, ``Toward generating a new intrusion detection dataset and intrusion traffic characterization,'' in {\em Proceedings of the International Conference on Information Systems Security and Privacy (ICISSP)}, vol.~1, pp.~108--116, {IEEE}, 2018.

\bibitem{8888419}
I.~Sharafaldin, A.~H. Lashkari, S.~Hakak, and A.~A. Ghorbani, ``Developing realistic distributed denial of service (ddos) attack dataset and taxonomy,'' in {\em 2019 International Carnahan Conference on Security Technology (ICCST)}, pp.~1--8, 2019.

\bibitem{Sutanto2024}
T.~Sutanto, M.~Aditya, H.~Budiman, M.~Ridha, U.~Syapotro, and N.~Azijah, ``Comparison of logistic regression, random forest, svm, knn algorithm for water quality classification based on contaminant parameters,'' {\em INTI Journal}, vol.~2022, 11 2024.

\bibitem{price2023}
J.~Price, T.~Yamazaki, K.~Fujihara, and H.~Sone, ``Xgboost and support vector machines: Comparing the interpretability of machine learning models,'' {\em SSRN Electronic Journal}, 01 2023.

\end{thebibliography}
\bibliographystyle{ieeetr}  

\end{document}